\documentclass[lettersize,journal]{IEEEtran}
\usepackage{amsmath,amsfonts}
\usepackage{array}
\usepackage{textcomp}
\usepackage{stfloats}
\usepackage{url}
\usepackage{verbatim}
\usepackage{breqn}

\usepackage{lineno,hyperref}
\modulolinenumbers[5]
\usepackage{amsmath,amsfonts}
\usepackage{amssymb}
\usepackage{amsthm}
\usepackage{array}
\usepackage{stfloats}

\usepackage{soul}
\usepackage{graphics}
\usepackage{graphicx}
\usepackage{subfigure}
\usepackage{epsfig}
\usepackage{epstopdf}
\usepackage{blkarray}
\usepackage{multirow}
\usepackage{booktabs}
\usepackage{float}
\usepackage{soul}
\usepackage{pdflscape}
\usepackage{mathtools}
\usepackage[linesnumbered,ruled,vlined]{algorithm2e}
\usepackage{qtree}
\usepackage{setspace}
\usepackage{color}

\newtheorem{mydef}{Definition}

\newtheorem{pro}{Property}[section]

\usepackage{graphicx}
\usepackage{cite}
\begin{document}

\title{Inter Observer Variability Assessment through Ordered Weighted Belief Divergence Measure in MAGDM: Application to the Ensemble Classifier Feature Fusion }

\author{\IEEEauthorblockN{Pragya Gupta, Student Member, IEEE Debjani Chakraborty, Debashree Guha, Senior Member, IEEE}
\thanks{Pragya Gupta and Debjani Chakraborty are with the Department of Mathematics, Indian Institute of Technology Kharagpur, Kharagpur 721302, India (e-mail: debjani@maths.iitkgp.ac.in, g.pragya1@gmail.com).\\
Debashree Guha is with the School of Medical Science and Technology, Indian Institute of Technology Kharagpur, Kharagpur 721302, India (e-mail: deb1711@gmail.com).}
}
\maketitle

\begin{abstract}
A large number of multi-attribute group decision-making (MAGDM) have been widely introduced to obtain consensus results. However, most of the methodologies ignore the conflict among the experts' opinions and only consider equal or variable priorities of them. Therefore, this study aims to propose an Evidential MAGDM method by assessing the inter-observational variability and handling uncertainty that emerges between the experts. The proposed framework has fourfold contributions. First, the basic probability assignment (BPA) generation method is introduced to consider the inherent characteristics of each alternative by computing the degree of belief. Second, the ordered weighted belief and plausibility measure is constructed to capture the overall intrinsic information of the alternative by assessing the inter-observational variability and addressing the conflicts emerging between the group of experts. An
ordered weighted belief divergence measure is constructed to acquire the weighted support for each group of experts to obtain the final preference relationship. Finally, we have shown an illustrative example of the proposed Evidential MAGDM framework. Further, we have analyzed the interpretation of Evidential MAGDM in the real-world application for ensemble classifier feature fusion to diagnose retinal disorders using optical coherence tomography images. 
\end{abstract}


\section{Introduction}{\label{sec 1}}

Multi-attribute group decision-making (MAGDM) is one of the most crucial parts of modern decision science, which includes the process of determining the best optimal solution from all multiple decision attributes and alternatives \cite{anderson2018introduction,huang2023automatic} with respect to the group of experts. The key role is to support the decision analysts to take all necessary objectives and subjective attributes of the problem into contemplation to employ rational and explicit decision procedure \cite{wang2022two,huang2023automatic}. MAGDM has been extensively applied to various fields such as medicine, management, pattern analysis domain, and classification \cite{cao2019multi,george2024two,al2024fundus,hussain2024human,huang2023automatic}, which has acquired more attention from researchers \cite{anderson2018introduction,huang2023automatic,ye2024preference}. In pattern classification, the MAGDM is generally used in feature fusion to acquire a desirable recognition interpretation based on the fused features. The MAGDM problems exhibit three key characteristics: alternatives, multiple attributes with incomparable units, and multiple experts, in which the expert's weights holds a 
significant part. Determining the weight of experts is an essential research area. Several techniques have been introduced for determining the weights of the experts \cite{yue2012approach,pang2024concept}. 
\\
The traditional group decision-making approaches commonly focus on calculating results through the majority and voting principles, which is a straightforward arithmetic aggregation procedure ignoring trade-offs among alternatives and experts. Due to that, group decision-making (GDM) approaches \cite{zhang2021personalized,dong2020consensus} have been proposed to overcome the inconsistency and compute the optimal agreed results. In this regard, some approaches \cite{french1956formal,theil1963symmetry, yue2011deriving,yue2012approach} used the relative importance or preference relations between the group of experts by determining the influence relations. The weight determination methods that are focused on pairwise comparisons include two classes of approaches \cite{forman1998aggregating} that can be utilized to intervene in the discrepancy between the experts. In the first approach, the individual decisions are aggregated, in which initially, the pairwise comparison of different experts is incorporated into one, and the resultant aggregation pairwise comparison is treated as a single expert problem 
\cite{forman1998aggregating,blagojevic2016heuristic}. However, in the other class, the individuals' preference is aggregated, in which a weight vector is generated for each expert \cite{jin2017weighting,cao2019multi,xiao2022generalized}. The generated set of weight vectors is transformed into a single weight vector by employing Dempster Shafer's theory or utilizing various types of aggregation operators to determine the optimal weights.
Although these approaches are simple to implement, but the information loss is extensive, and while computing the weights for experts, the inter-observational variability among the alternatives is not taken into consideration. These aforementioned challenges motivate us to propose a MAGDM approach that considers the degree of belief for each alternative which includes the inter-observational variability between the group of experts. 
\\
With this view, this study aims to develop a MAGDM framework based on the Evidential ordered weighted belief divergence measure to evaluate the support for each expert and transform it into a consensus situation. For computing the ordered weighted belief divergence measure, the construction of ordered weighted belief and plausibility is proposed. The basic probability assignment (BPA) generation method has been introduced to evaluate the ordered weighted belief and plausibility measure, which analyzes the crucial information by considering the alternative responses corresponding to the attributes. It helps to assess the belongingness of each alternative corresponding to the domain of the attribute that is partitioned by selecting the suitable number of linguistic term sets followed by computing BPAs. The empirical study of the proposed Evidential MAGDM technique is demonstrated over an illustrative example to show the efficacy of the proposed approach. We have shown the effectiveness of the proposed technique in real-world problems to assess the capability of handling uncertainty and inter-observational behavior in the feature fusion module for the diagnosis of retinal disorders. In our study, we propose an ensemble classifier feature fusion using the proposed Evidential MAGDM approach for the classification of various retinal disorders by integrating the multi-expert view phenomena. The proposed system includes three expert paths to generate the OCT dataset at a different scale space followed by integrating the EfficientNet-B0 model \cite{tan2019efficientnet}. 
The model extracts the discriminative feature representation at various scales to capture the diversity in feature representation locally and globally, which influences the classification probability for the diagnosis of retinal disease through OCT images. The extracted features from each path are fused together using the proposed Evidential MAGDM approach by handling the conflict and impreciseness emerging between the distinctive features at various scales, followed by the classification of retinal disorders. 
In a nutshell, the contribution to this is recapitulated as follows: 
\begin{itemize}
    \item This study proposes a generation framework of BPAs for MAGDM with respect to the group of experts by computing the belief degree based on partitioning the domain of attribute into linguistic terms.
    \item We propose an ordered weighted belief measure to compute the overall belongingness by considering inter-observational variability evidence of each alternative corresponding to the attributes and to analyze the conflicts and discrepancy between the group of experts; the ordered weighted plausibility measure is proposed.
    \item We propose an ordered weighted belief generalized divergence measure to compute the weighted support for each expert for MAGDM.
    \item We propose an Evidential MAGDM framework based on the constructed ordered weighted belief divergence measure to compute overall weights for each group of experts.
    \item To analyze the effectiveness of the proposed Evidential MAGDM approach, an illustrative example is demonstrated.
    \item We propose the application of the Evidential MAGDM method in the ensemble classifier feature fusion for the diagnosis of retinal disorders. 
\end{itemize}
The remainder of this manuscript is organized as follows: Section \ref{sec 2} briefly introduces the key concepts of BPA and divergence measures. Section \ref{sec 3} introduces the Evidential MAGDM approach.
In Section \ref{sec 4}, a numerical example and comparative analysis are provided to validate the effectiveness of the proposed approach, followed by an application to the ensemble classifier feature fusion for diagnosing the retinal disorder. Finally, concluding remarks are given in Section \ref{sec 5}.  

\section{Prerequisites}{\label{sec 2}}
\subsection{Theory of Belief Functions}{\label{sec 2.1}}
Various approaches have been exploited to address the uncertainty in recent times. DS evidence theory of the belief function is one of the tools that provides a general technique for modeling uncertainty and is widely executed in various areas. This section proceeds towards presenting the basic concepts that are considered helpful in the subsequent work. 

\begin{mydef} {Frame of Discernment} {\cite{dempster2008upper}}:{\label{def 1}}
  Let $\Omega$ be a set of $n$ mutually exclusive elements, which can be expressed as $\Omega = \{\omega_{1},...,\omega_{n}\}$. Then, the set $\Omega$ is called a frame of discernment. The power set of $\Omega$ is defined as $2^{\Omega}=\{\varnothing,\{\omega_{1}\},\{\omega_{1},\omega_{2}\},...,\Omega\}$. If $U\in 2^{\Omega}$, $U$ is called proposition.
\end{mydef}

\begin{mydef} {Basic Probability Assignment} {\cite{dempster2008upper,shafer1976mathematical}}:{\label{def 2}}
    It is a mapping $m:2^{\Omega}\mapsto [0,1]$, which satisfies:
    \begin{equation}{\label{eq 1}}
        m(\varnothing)=0, \quad \sum_{U\subseteq \Omega} m(U)=1
    \end{equation}
If $m(U)>0$, then $U$ is said to be a focal element. The value $m(U)$ signifies the belief degree to support proposition $U$. In DS evidence theory, $m$ is also called a basic belief assignment (BBA). 
\end{mydef}
In DS evidence theory, the belief and plausibility function is related to BPA, which represents the lower and upper bounds for each proposition in a BPA, and it can be described as follows:

\begin{mydef} {Belief Function} {\cite{shafer1976mathematical}}:{\label{def 3}}
    It is a mapping $Bel:2^{\Omega}\mapsto [0,1]$ and it can be defined as:
    \begin{equation}{\label{eq 2}}
      Bel(U)=\sum_{A\subseteq U}m(A)    
    \end{equation}
\end{mydef}

\begin{mydef} {Plausibility Function} {\cite{shafer1976mathematical}}:{\label{def 4}}
    It is a mapping $Pl:2^{\Omega}\mapsto [0,1]$, and it can be defined as:
    \begin{equation}{\label{eq 3}}
        Pl(U)=\sum_{A\cap U \neq \varnothing}m(U), \hspace{0.1 cm} Pl(U)=1-Bel(\overline{U})
    \end{equation}
    
\end{mydef}
It is to be noted that $Pl(U)\geq Bel(U)$ for each $U \subseteq \Omega$.
\begin{mydef} {Dempster's Combination Rule} {\cite{dempster2008upper}}:{\label{def 5}}
Let two BPAs $m_{1}$ and $m_{2}$ that are independent in frame of discernment $\Omega$, Dempster's rule of combination $m_{1} \oplus m_{2}$, is defined as follows:
\begin{equation}{\label{eq 4}}
    (m_{1} \oplus m_{2})A= \left\{ 
  \begin{array}{ c l }
    \frac{1}{1-K}\sum_{\substack{A_{1}\cap A_{2}=A}}m_{1}(A_{1})m_{2}(A_{2}), & A \neq \varnothing  \\
    0,                 &  A=\varnothing
  \end{array}
\right.
\end{equation}
with $K=\sum_{A_{1}\cap A_{2}=\varnothing} m_{1}(A_{1})m_{2}(A_{2}) $, where $K$ denotes the coefficient of conflicts between BPAs $m_{1}$ and $m_{2}$. 
\end{mydef}
\subsection{Generalized Jensen Shannon Divergence Measures}{\label{sec 2.2}}
\raggedbottom
In information theory, divergence measures are employed to measure the discrepancy of data information and are widely exploited in various application areas \cite{he2003generalized,xiao2022generalized}. The Kullback-Leibler (KL) divergence \cite{kullback1997information} is one of the popular divergence measures which is non-negative, additive, but not symmetric. In this regard, a new divergence measure called the Jensen-Shannon divergence measure \cite{lin1991divergence} is introduced such that it is symmetric but also satisfies the boundedness and the range between 0 to 1 for the logarithmic base 2, which is illustrated below.
\begin{mydef}{\cite{lin1991divergence}}{\label{def 7}}
    The Jensen-Shannon divergence between two probability distributions $A=\{a_{1},a_{2},...,a_{n}\}$ and $B=\{b_{1},b_{2},...,b_{n}\}$ is defined as follows:
    \begin{equation}{\label{eq 6}}
        JS(A,B)=\dfrac{1}{2}\Big[I(A,\dfrac{A+B}{2})+I(B,\dfrac{A+B}{2})\Big]
    \end{equation}
    where $I(A,B)$ is the KL divergence \cite{kullback1997information}.
\end{mydef}

\begin{mydef}{Generalized Jensen-Shannon Divergence}{\cite{lin1991divergence}}:{\label{def 7*}}
    Let $A_{1},A_{2},...,A_{p}$ be $p$ probability distributions with weights $w_{1},w_{2},...,w_{p}$, respectively. The generalized Jensen-Shannon (JS) divergence can be expressed as
    \begin{equation}{\label{eq 6*}}
        JS_{w}(A_{1},A_{2},...,A_{p})=H(\sum_{i=1}^{p}w_{i}A_{i})-\sum_{i=1}^{p}w_{i}H(A_{i})
    \end{equation}
    where $H(A_{i})=-\sum_{i}a_{ji}loga_{ji}$ with $\sum a_{ij}=1$ $(i=1,2,...,n; j=1,2,...,p)$.
\end{mydef}
Belief Jensen-Shannon (BJS) divergence measure was proposed in \cite{wang2021new} to measure the conflicts between two BPAs by considering the belief and plausibility measure and it can be defined as:

\begin{mydef}{\cite{wang2021new}}{\label{def 8}}
    The divergence measure between two independent BPAs $m_{1}$ and $m_{2}$ define on $\Omega$ is defined as:
    \begin{equation}{\label{eq 7}}
       \begin{aligned}
           Div(m_{1},m_{2})=\dfrac{1}{2}\Big[I(WPBl_{m_{1}},\dfrac{WPBl_{m_{1}}+WPBl_{m_{2}}}{2}) \\
        +I(WPBl_{m_{2}},\dfrac{WPBl_{m_{1}}+WPBl_{m_{2}}}{2})\Big]
       \end{aligned}
    \end{equation}
    where, $WPBl_{m}(A_{i})=\dfrac{Bel(A_{i})+Pl(A_{i})}{\sum_{A_{i}\subseteq \Omega}Bel(A_{i})+Pl(A_{i})}$.
\end{mydef}

The $Div(m_{1},m_{2})$ can also be expressed in the following form:
\begin{equation}{\label{eq 8}}
    \begin{aligned}
    Div(m_{1},m_{2})= 
    \\
    \dfrac{1}{2}\Big[\sum_{A_{i}\subseteq \Omega} WPBl_{m_{1}}(A_{i})\log \dfrac{2WPBl_{m_{1}}(A_{i})}{WPBl_{m_{1}}(A_{i})+WPBl_{m_{2}}(A_{i})} + \\
    \sum_{A_{i}\subseteq \Omega}WPBl_{m_{2}}(A_{i})\log \dfrac{2WPBl_{m_{2}}(A_{i})}{WPBl_{m_{1}}(A_{i})+WPBl_{m_{2}}(A_{i})} \Big]
    \end{aligned}
\end{equation}
Next, we propose a BPA generation method to evaluate the degree of support for each alternative with respect to the attributes for group decision-making, followed by the construction of ordered weighted belief and plausibility function. To calculate the weighted support for each expert for the fusion, a novel construction method for a generalized weighted belief divergence measure is proposed, which will consider the inter-observational variability between the alternatives and handle the conflicts occurring between the group of experts.  

\section{The proposed Evidential MAGDM methodology}{\label{sec 3}}
This section introduces a BPA generation method for each alternative corresponding to the attribute associated with experts using the membership function for the linguistic value, followed by the construction of ordered weighted belief and plausibility function. To handle the conflicts occurring between the group of experts and impreciseness, a generalized ordered weighted belief divergence measure is proposed. 

\subsection{BPAs generation for a group decision making}{\label{sec 3.1}}

\begin{figure}[htp]
    \centering
    \includegraphics[width=2.9in]{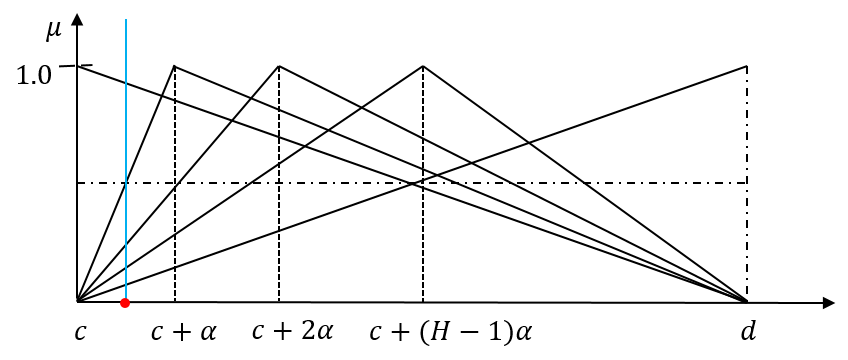}
    \caption{Linguistic values membership function with $(H+1)$ terms}
    \label{fig 1}
\end{figure}

In this study, we assume that $\{A_{1},A_{2},...,A_{p}\}$ $(p\geq 2, P=\{1,2,...,p\})$ be a discrete set of feasible alternatives, $\{t_{1},t_{2},...,t_{q}\}$ $(Q=\{1,2,...,q\})$ be a set of attributes; $\{u_{1},u_{2},...,u_{k}\}$ $(K=\{1,2,...,k\})$ be the group of experts. A multi-attribute group decision-making (MAGDM) can be represented as follows:
\begin{equation}{\label{eq 9}}
    Y_{k} = \Big(y_{ij}^{(k)}\Big)_{p\times q} =\bordermatrix{ & t_{1} & t_{2} & \ldots & t_{q} \cr
              A_{1} & y_{11}^{(k)} & y_{12}^{(k)} & \ldots &y_{1q}^{(k)} \cr
              A_{2} & y_{21}^{(k)} & y_{22}^{(k)} & \ldots &y_{2q}^{(k)} \cr
              \vdots  & \vdots & \vdots & \ldots & \vdots \cr 
              A_{p} & y_{p1}^{(k)} & y_{p2}^{(k)} & \ldots & y_{pq}^{(k)}}, \quad k\in K
\end{equation}
be the decision matrix provided by $k$th expert. In order to evaluate all attributes in dimensionless units and facilitate inter-attribute comparisons, each attribute of the considered decision matrix is normalized \cite{yue2011method} using Eqs. \ref{eq 10*}
and obtain corresponding element $y_{ij}^{(k)}$ in normalized decision matrix $Y_{k}$.
\begin{equation}{\label{eq 10*}}
    y_{ij}^{(k)}=\dfrac{y_{ij}^{(k)}}{\sqrt{\sum_{i=1}^{p}(y_{ij}^{(k)})^{2}}} \quad (i\in P,j \in Q,k \in K),  
\end{equation}
for benefit attribute $t_{j}$.\\
As considered above, there are $k$ experts; each expert provides their preference over the alternatives corresponding to attributes. These provided subjective/objective preferences for the alternative may differ according to the group of experts, which influences the final decision-making framework. The primary goal is to compute the weighted support for the above-considered MAGDM problem by considering this fact.\\
Due to subjectiviy in experts' opinions, it is assumed that there is still uncertainty in their expressions. Opinions are not fully precise. In a broader sense, it may be said that there is a chance of variation of a particular opinion for some external causes. Causes may be - expert's attitude, peer's influence, incomplete knowledge, the complexity of studied systems, or any other environmental factors. In order to capture these chance factors, we present the framework to compute the degree of belief in terms of basic probability assignment (BPA). Another challenge is to capture the diversity emerging among the group of experts without hindering the degree of belief (support) associated with the alternatives.  Consequently, we propose a weighted belief divergence measure to assess the diversity emerging between the group of experts for each alternative. It is computed based on the ordered weighted belief and plausibility measure, which handles the conflicts occurring between the group of experts and comprises the degree of belief for each alternative. The ordered weighted divergence measure relies on belief and plausibility measures, which capture the degree of confidence by evaluating the support of belongingness in the domain of the attributes, followed by examining the provided subjective/objective preference relationship among the experts for each alternative to handle the conflicting information. The subjective preference of the alternative is stretched over the domain of attribute for the evaluation of the ordered weighted belief and plausibility measure, which analyzes the inter-observational variability between the group of experts. For computing belief and plausibility measures, we need to evaluate basic probability assignment, which defines the degree of confidence. For computing the degree of belief for each alternative over the attribute domain, we have scaled the subjective preferences of alternatives over the attribute domain by employing the idea presented in \cite{chakraborty2016multi}. We have utilized the augmented number of linguistic terms for partitioning the domain of the attribute to capture the intrinsic characteristic and inter-observational variability among the alternatives for each linguistic term set, as shown in Fig. \ref{fig 1}. 
In this work, it is assumed that each linguistic variable $y_{1},y_{2},...,y_{p}$ values are defined in the interval $[c,d]\subset \mathbb{R}$ (where, $[c,d]$ represents the attribute's domain). Consider that $Y=[c,d]$ and $\Im (y)$ consists of $H+1$, $(H\geq 2)$, terms shown in Fig. \ref{fig 1} as elaborated below.\\
$\Im = \{low_{1},around(c+\alpha),around(c+2\alpha),...,around(c+(H-1)\alpha),high_{H}\}$;\\
where $\alpha=\dfrac{d-c}{H}$, and each term can be represented using the triangular membership functions $\{\mu_{\tilde{B_{1}}},\mu_{\tilde{B_{2}}},...,\mu_{\tilde{B_{H+1}}}\}$ of the subsequent form:
\begin{equation}{\label{eq 11}}
    \mu_{\tilde{B_{1}}}(y) = \mu_{low_{1}}(y)
    = \left\{ 
  \begin{array}{ c l }
    1-\dfrac{(y-c)}{(d-c)} & \quad if \quad  c \leq y \leq d   \\
    0                 & \quad otherwise,
  \end{array}
\right.
\end{equation}

\begin{dmath}{\label{eq 12}}
    \mu_{\tilde{B}_{h}}(y) = \mu_{around(c+h\alpha)}(y) 
    = \left\{ 
  \begin{array}{ c l }
    1-\dfrac{(c+h\alpha-y)}{h\alpha} & \quad if \quad  c \leq y \leq c+h\alpha   \\
    1-\dfrac{(y-c-h\alpha)}{(d-c-h\alpha)} & \quad if \quad c+h\alpha \leq y \leq d \\ 
    0                 & \quad otherwise
  \end{array}
\right.
\end{dmath}

where $1 \leq h \leq (H-1)$.

\begin{dmath}{\label{eq 13}}
    \mu_{\tilde{B}_{H+1}}(y) = \mu_{high_{H}}(y) \\
    = \left\{ 
  \begin{array}{ c l }
    1-\dfrac{(d-y)}{(d-c)} & \quad if \quad  c \leq y \leq d   \\
    0                 & \quad otherwise,
  \end{array}
\right.
\end{dmath}

In this work, we have determined a $l$ linguistic term set associated with triangular membership functions
$\{\mu_{\tilde{B_{1}}},\mu_{\tilde{B_{2}}},...,\mu_{\tilde{B_{l}}}\}$ to characterize the alternative's belongingness with respect to the attributes.
\begin{equation}{\label{eq 14}}
    \begin{aligned}
    R_{k} =\\
    \bordermatrix{ &  &  &  &  & & & &  \cr
                & \mu_{t_{1}\Tilde{B}_{1}}^{k_{A_{1}}} & 
                \ldots &
                \mu_{t_{1}\Tilde{B}_{l}}^{k_{A_{1}}} & \vdots & \ldots & \vdots & \mu_{t_{q}\Tilde{B}_{1}}^{k_{A_{1}}} & 
                \ldots &
                \mu_{t_{q}\Tilde{B}_{l}}^{k_{A_{1}}} \cr
                & \mu_{t_{1}\Tilde{B}_{1}}^{k_{A_{2}}} & 
                \ldots &
                \mu_{t_{1}\Tilde{B}_{l}}^{k_{A_{2}}} & \vdots & \ldots & \vdots & \mu_{t_{q}\Tilde{B}_{1}}^{k_{A_{2}}} & 
                \ldots &
                \mu_{t_{q}\Tilde{B}_{l}}^{k_{A_{2}}} \cr
                & \ldots & \ldots & \ldots & \vdots & \ldots & \vdots & \ldots & \ldots & \ldots \cr 
                & \mu_{t_{1}\Tilde{B}_{1}}^{k_{A_{p}}} & 
                \ldots &
                \mu_{t_{1}\Tilde{B}_{l}}^{k_{A_{p}}} & \vdots & \ldots & \vdots & \mu_{t_{q}\Tilde{B}_{1}}^{k_{A_{p}}} & 
                \ldots &
                \mu_{t_{q}\Tilde{B}_{l}}^{k_{A_{p}}}
                },   
                \end{aligned}
\end{equation}
where $c = min_{1 \leq i \leq p }\{y_{ij}^{(k)}\}$, $d = max_{1 \leq i \leq p }\{y_{ij}^{(k)}\}$ $(k\in K)$ with respect to the attribute $t_{j}$, and $\mu^{m_{A_{i}}}_{t_{j}\Tilde{B}_{l}}$, $(i \in P, \quad j \in Q, \quad k\in K$)  
represents the degree of belongingness in the partitioned domain $\Tilde{B}_{l}$ for the attribute $t_{j}$ corresponding to the $k$th expert. To obtain the basic probability assignments (BPAs) for each alternative corresponding to the attributes, the above matrix is normalized column-wise as follows to satisfy the criterion for the BPAs given in Eq. \ref{eq 1}.
    \begin{equation}{\label{eq 15}}
    \begin{aligned}
     \Bar{R}_{k} =\\
    \bordermatrix{ &  &  &  &  & & & &  \cr
                & m_{t_{1}\Tilde{B}_{1}}^{k_{A_{1}}} & \ldots  & 
                m_{t_{1}\Tilde{B}_{l}}^{k_{A_{1}}} & \vdots & \ldots & \vdots & m_{t_{q}\Tilde{B}_{1}}^{k_{A_{1}}} & \ldots & m_{t_{q}\Tilde{B}_{l}}^{k_{A_{1}}} \cr
                & m_{t_{1}\Tilde{B}_{1}}^{k_{A_{2}}} & \ldots & m_{t_{1}\Tilde{B}_{l}}^{k_{A_{2}}} & \vdots & \ldots & \vdots & m_{t_{q}\Tilde{B}_{1}}^{k_{A_{2}}} & \ldots & m_{t_{q}\Tilde{B}_{l}}^{k_{A_{2}}} \cr
                & \ldots & \ldots & \ldots & \vdots & \ldots & \vdots & \ldots & \ldots & \ldots  \cr 
                & m_{t_{1}\Tilde{B}_{1}}^{k_{A_{p}}} & \ldots & m_{t_{1}\Tilde{B}_{l}}^{k_{A_{p}}} & \vdots & \ldots & \vdots & m_{t_{q}\Tilde{B}_{1}}^{k_{A_{p}}} & \ldots & m_{t_{q}\Tilde{B}_{l}}^{k_{A_{p}}}     
                 }, 
    \end{aligned}
\end{equation}

where 
$m_{t_{j}\Tilde{B}_{l}}^{k_{A_{i}}}=\dfrac{\mu_{t_{j}\Tilde{B}_{l}}^{k_{A_{i}}}}{\sum_{i=1}^{p}\mu_{t_{j}\Tilde{B}_{l}}^{k_{A_{i}}}}$ 
$(i\in P, j\in Q, k \in K)$.
\\
The decision matrix $\Tilde{R}_{k}$ in Eq. \ref{eq 15} represents the BPAs of each alternative with respect to the attributes corresponding to the $k$th expert, which indicates the degree of support (belief) with respect to each partitioned region to handles the impreciseness between the alternative data information. Next, we construct the ordered weighted belief and plausibility function to evaluate the overall support for each alternative by assessing the conflicts occurring in the decisions.
\subsection{Construction of Ordered Weighted Belief, Plausibility, and Ordered Weighted Belief Divergence Measure based on experts}{\label{sec 3.2}}
This section introduces the ordered weighted belief function and plausibility function, followed by a generalized ordered weighted divergence measure based on the belief and plausibility function.

\begin{mydef}{\label{def 9}}
 The ordered weighted belief function based on the weighting vector $w = (w_{1},w_{2},...,w_{l})$, with $\sum_{f=1}^{l}w_{f}=1$, $w_{f}\in [0,1]$, for alternative $A_{i}$, with respect to the attribute $t_{j}$ is defined as:\\
 \begin{equation}{\label{eq 16}}
  Bel_{wk}^{t_{j}}(A_{i}) = \sum_{f=1}^{l}w_{f}m_{t_{j}\Tilde{B}_{\sigma{(f)}}}^{k_{A_{i}}}, \quad i\in P,j\in Q 
 \end{equation}
 where $\sigma:\{1,2,...,l\} \mapsto \{1,2,...,l\}$ is a permutation such that $m_{t_{j}\Tilde{B}_{\sigma{(1)}}}^{k_{A_{i}}} \geq m_{t_{j}\Tilde{B}_{\sigma{(2)}}}^{k_{A_{i}}} \geq ... \geq m_{t_{j}\Tilde{B}_{\sigma{(l)}}}^{k_{A_{i}}}$.
\end{mydef}

Here, $Bel_{wk}^{t_{j}}(A_{i})$ represents a total amount of belief supporting necessarily to the alternative $A_{i}$ corresponding to the attribute $t_{j}$ with respect to the $k$the expert by giving the higher weight to the most supported belief degree using the orness concept for weight determination \cite{o1988aggregating}. The more dispersed the $w$, the more individual belief degree is used to evaluate the overall support for alternative $A_{i}$ with respect to the $k$th expert. 

\begin{mydef}{\label{def 10}}
    The plausibility function for alternative $A_{i}$ based on the expert $u_{k}$ corresponding to the attribute $t_{j}$ is defined as  
    \begin{equation}{\label{eq 17}}
    Pl_{wk}^{t_{j}}(A_{i})=1-\dfrac{\sum_{\substack{i=1 \\ i \neq k}}^{k} Bel_{wi}^{t_{j}}(A_{i})}{\sum_{i=1}^{k} Bel_{wi}^{t_{j}}(A_{i})}    
    \end{equation}
\end{mydef}
Where $Pl_{wk}^{t_{j}}(A_{i})$ quantifies the overall amount of belief for alternative $A_{i}$ with respect to the group of experts by enduring the conflicts and impreciseness emerging between the experts. Next, we define an ordered weighted belief divergence measure based on the ordered weighted belief and plausibility function to measure discrepancy and conflict among the group of experts by considering the inter-observational variability between the group of experts for the alternatives.

\begin{mydef}{\label{def 11}}
    Given two independent BPAs $m_{1}$ and $m_{2}$, define on the frame of discriminant $\Omega$, the  weighted divergence between $m_{1}$ and $m_{2}$ based on set of weights $w_{1},w_{2}\geq 0$ with $w_{1}+w_{2}=1$ is defined as:
    \begin{equation}{\label{eq 20}}
        \begin{aligned}
            Div_{w}(m_{1},m_{2})=
         \sum_{i=1}^{2}w_{i}\sum_{A_{j}\subseteq \Omega}WPBl_{m_{\sigma(i)}}(A_{j}) \\
        \log\dfrac{WPBl_{m_{\sigma (i)}}(A_{j})}{w_{1}WPBl_{m_{\sigma(1)}}(A_{j})+w_{2}WPBl_{m_{\sigma(2)}}(A_{j})}
        \end{aligned}
    \end{equation}
\end{mydef}
When $w_{i} = \dfrac{1}{2}$ $(i \in \{1,2\})$ and $WPBl_{m_{\sigma(1)}}(A_{j}) \geq WPBl_{m_{\sigma(2)}}(A_{j})$, then the Eq. \ref{eq 20} can be expressed in Eq. \ref{eq 7}. The Eq. \ref{eq 20} can be expressed as the Jensen-Shannon (JS) divergence measure, which depends on the weights \cite{lin1991divergence}. One of the key features of the weighted JS divergence is that we can allocate distinct weights to the distributions implicated according to their prominence. Therefore, the weights associated in Eq. \ref{eq 20} restrain the trade-off behavior arising between the data information with respect to the group of experts in the evaluation of the divergence measure. To consider the discrepancy between groups of experts, the generalized weighted belief divergence based on Jensen Shannon inspired from \cite{xiao2022gejs} in terms of belief and plausibility for multiple BPAs is defined as follows: 

\begin{mydef}{\label{def 12}}
    Given the set of BPAs $\{m_{1},m_{2},...,m_{p}\}$ over the frame of discriminant $\Omega$ with corresponding set of weights $\{w_{1},...,w_{p}\}$ with $\sum_{i=1}^{p}w_{i}=1$, the generalized weighted divergence is defined as:
    \begin{equation}{\label{eq 21}}
        \begin{aligned}
         GDiv_{w}(m_{1},m_{2},...,m{p})= 
        \sum_{i=1}^{p}w_{i}\sum_{A_{j}\subseteq \Omega}\dfrac{WPBl_{m_{\sigma(i)}}(A_{j})}{|A_{j}|} \\
        \log\dfrac{WPBl_{m_{\sigma(i)}}(A_{j})}{w_{1}WPBl_{m_{\sigma(1)}}(A_{j})+...+w_{p}WPBl_{m_{\sigma(p)}}(A_{j})}
    \end{aligned}
    \end{equation}
\end{mydef}
where, $WPBl_{m_{(\sigma(1))}}(A_{j}) \geq ... \geq WPBl_{m_{(\sigma(p))}}(A_{j})$ for any permutation $\sigma : \{1,...,p\} \mapsto \{1,...,p\}$.
When $p=2$ and $|A_{j}|=1$, then the Eq. \ref{eq 21} is reduced in the Eq. \ref{eq 20}.
Next, we present the properties of the proposed generalized divergence measure.
\begin{pro}{\label{property 1}}
    $GDiv_{w}(m_{1},...,m_{i},...,m_{p})$ is bounded, where $0 \leq GDiv_{w}(m_{1},...,m_{i},...,m_{p}) \leq \log p$.
\end{pro}

\begin{pro}{\label{property 2}}
    $GDiv_{w}(m_{1},...,m_{i},...,m_{p})=0$ if and only if $m_{i}$ $(1\leq i\leq p)$ are equal. 
\end{pro}

\begin{pro}{\label{property 3}}
The proposed divergence measure $GDiv_{w}$ is symmetrical, $GDiv_{w}(m_{1},...,m_{p})=GDiv_{w}(m_{\sigma(1)},...,m_{\sigma(p)})$ for any permutation $\sigma:\{1,...,p\}\mapsto \{1,...,p\}.$   
\end{pro}

Next, we presented the framework of the Evidential MAGDM technique by computing the belief degree for each alternative for determining the expert weight to obtain the agreed results by considering all the necessary aspects of the group of experts.
\subsection{Proposed framework for group decision making}
\begin{figure*}[htp]
    \centering
    \includegraphics[width=4.8in]{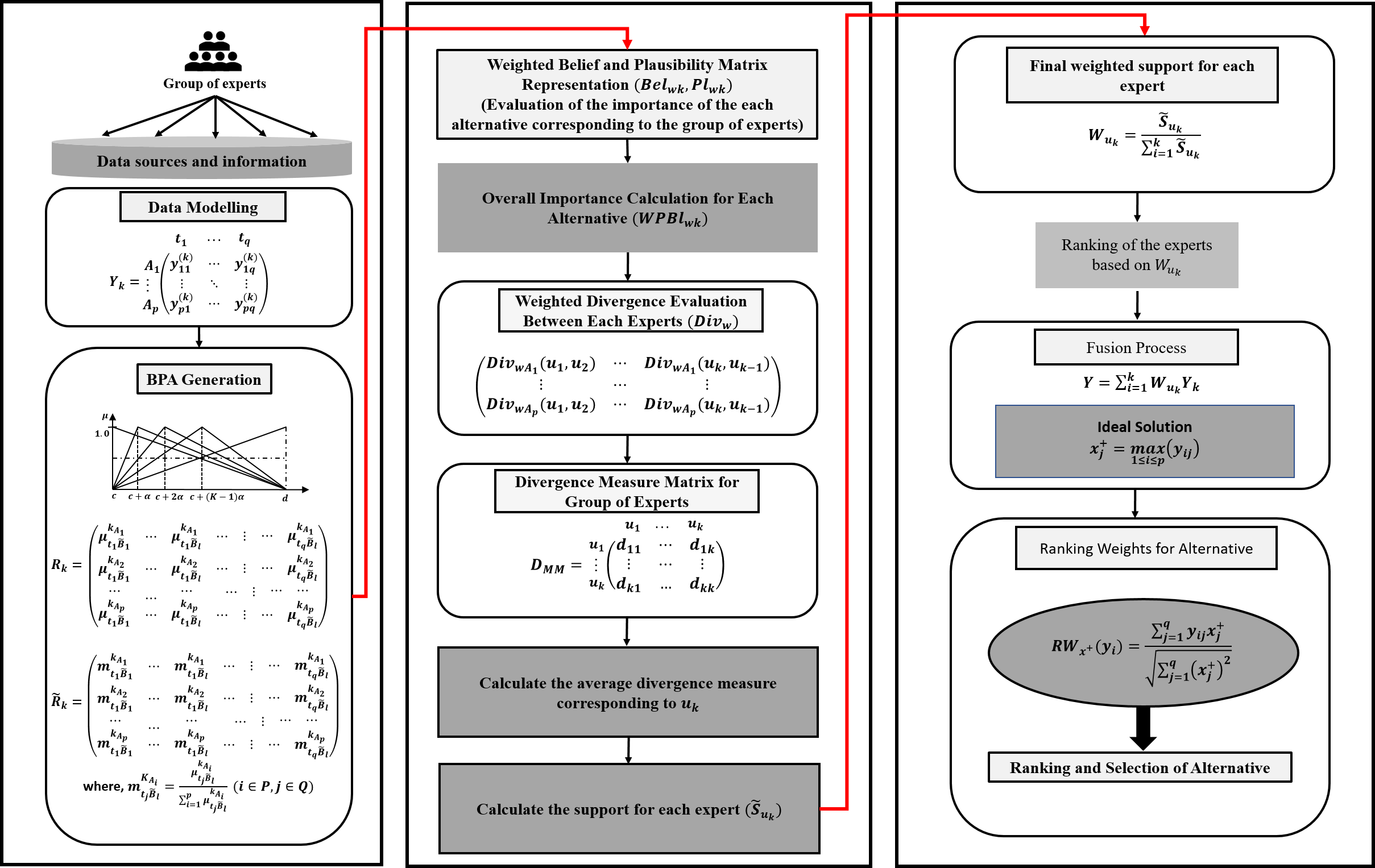}
    \caption{The proposed framework for group decision-making by analyzing the inter-observational variability between each alternative corresponding to the attributes for handling the impreciseness and conflicts between the group of experts.}
    \label{fig 2}
\end{figure*}

{\label{sec 3.3}}
This section introduces the proposed framework for group decision-making demonstrated in Fig. \ref{fig 2} by conceptualizing the novel ordered weighted belief and plausibility functions followed by the ordered weighted belief divergence measure for computing the expert's weight by considering inter-observational variability and controlling the impact of conflicting information, which will not affect the final decision. The proposed framework for MAGDM for evaluating an expert's weight can be elaborated in the following steps:\\  
\textbf{Step 1.} Generate BPAs for the normalized decision matrix of Eq. \ref{eq 9} based on the proposed method for computing BPAs for each alternative corresponding to each expert presented in section \ref{sec 3.1}.\\
\textbf{Step 2.} Determine the ordered weighted belief measure for $k$th expert using Eq. \ref{eq 16}.

\begin{equation}{\label{eq 22}}
    Bel_{wk} =\bordermatrix{ &  &  &  &  \cr
               & Bel_{wk}^{t_{1}}(A_{1}) & Bel_{wk}^{t_{2}}(A_{1}) & \ldots & Bel_{wk}^{t_{q}}(A_{1}) \cr
               \cr
               & Bel_{wk}^{t_{1}}(A_{2}) & Bel_{wk}^{t_{2}}(A_{2}) & \ldots & Bel_{wk}^{t_{2}}(A_{2}) \cr
               & \vdots & \vdots & \ldots & \vdots \cr 
               & Bel_{wk}^{t_{1}}(A_{p}) & Bel_{wk}^{t_{2}}(A_{p}) & \ldots &  Bel_{wk}^{t_{2}}(A_{p})},
\end{equation}

\textbf{Step 3.} Determine the weighted plausibility measure for $k$th expert using Eq. \ref{eq 17}.
\begin{equation}{\label{eq 23}}
    Pl_{wk} =\bordermatrix{ &  &  &  &  \cr
               & Pl_{wk}^{t_{1}}(A_{1}) & Pl_{wk}^{t_{2}}(A_{1}) & \ldots & Pl_{wk}^{t_{q}}(A_{1}) \cr
               \cr
               & Pl_{wk}^{t_{1}}(A_{2}) & Pl_{wk}^{t_{2}}(A_{2}) & \ldots & Pl_{wk}^{t_{2}}(A_{2}) \cr
               & \vdots & \vdots & \ldots & \vdots \cr 
               & Pl_{wk}^{t_{1}}(A_{p}) & Pl_{wk}^{t_{2}}(A_{p}) & \ldots &  Pl_{wk}^{t_{2}}(A_{p})}
\end{equation}
\newline
\textbf{Step 4.} Calculate the weighted divergence measure by taking pair of experts with respect to the alternative in the following way using Eq. \ref{eq 20}.
\begin{equation}{\label{eq 25}}
    D =\bordermatrix{ &  &  &  &  \cr
               & Div_{wA_{1}}(u_{1},u_{2}) &  \ldots & Div_{wA_{1}}(u_{k},u_{k-1}) \cr
               \cr
               & Div_{wA_{2}}(u_{1},u_{2}) &  \ldots & Div_{wA_{2}}(u_{k},u_{k-1}) \cr
               & \vdots & \vdots &  \vdots \cr 
               & Div_{wA_{p}}(u_{1},u_{2}) &  \ldots & Div_{wA_{p}}(u_{k},u_{k-1}) \cr}
\end{equation}
\textbf{Step 5.} Determine the divergence measure matrix for the group of experts.
\begin{equation}{\label{eq 26}}
    D_{MM} =\bordermatrix{ & u_{1}  & u_{2} & \ldots  & u_{k} \cr
              u_{1} & d_{11} & d_{12} & \ldots & d_{1k} \cr
               \cr
               u_{2} & d_{21} & d_{22} & \ldots & d_{2k} \cr
               \vdots & \vdots & \vdots & \ldots & \vdots \cr 
               u_{k} & d_{k1} & d_{k2} & \ldots & d_{kk} \cr}
\end{equation}
where $d_{k_{1}k_{2}} = \sum_{i=1}^{p}Div_{wA_{i}}(u_{k_{1}},u_{k_{2}})$ and $k_{1},k_{2} \in K$.
\newline
\textbf{Step 6.} Calculate the average divergence measure for each experts using $D_{MM}$ matrix.
\begin{equation}{\label{eq 27}}
    d_{u_{j}} = \sum_{i=1}^{k}d_{ij}, \quad j \in K
\end{equation}
\newline
\textbf{Step 7.} Determine the support for each expert as follows using the average divergence measure illustrated in Step 6.
\begin{equation}{\label{eq 28}}
    \tilde{S}_{u_{k}} = \dfrac{1}{d_{u_{j}}}, \quad j\in K 
\end{equation}
\textbf{Step 8.} Calculate the final weighted support for each expert for group decision-making.
\begin{equation}{\label{eq 29}}
    W_{u_{k}} = \dfrac{\tilde{S}_{u_{k}}}{\sum_{i}^{k}\tilde{S}_{u_{k}}}
\end{equation}
\newline
\textbf{Step 9.} Computing the fusion of the attributes corresponding to the experts by providing the weight $W_{u_{k}}$ determined by previous steps in the following way.
\begin{equation}{\label{eq 30}}
    Y = \sum_{i}^{k}W_{u_{k}}Y_{k}
\end{equation}
\newline
\textbf{Step 10.} Ranking the alternatives in accordance with higher weighted response using \cite{yue2012approach} based on the ideal solution $x_{j}^{+} =max_{1 \leq i \leq p}(y_{ij})$ and it can be evaluated \cite{jahanshahloo2006algorithmic} as $RW_{x^{+}}(y_{i})=\dfrac{\sum_{j=1}^{q}y_{ij}x^{+}_{j}}{\sqrt{\sum_{j=1}^{q}(x_{j}^{+})^{2}}}$.
Next, we present an illustrative example of the simulated data using the proposed framework for MAGDM.
\section{Experimental analysis}{\label{sec 4}}
\subsection{Illustrative example}{\label{sec 4.1}}
In this section, a numerical example (adapted from \cite{shih2007extension}) is provided to illustrate the proposed approach. 
A company is endeavoring to recruit a manager. The human resources division of the company equips some applicable selection examinations as the benefit attributes to be assessed. These objective examinations contain knowledge and skill examinations. After these objective examinations, there are 17 eligible candidates on the inventory for selection. Then, four experts are reliable for the selection of the candidate, among others, based on subjective examinations. The primary data of subjective attributes, consisting of panel interviews and 1-on-1 interview examinations for the decisions, are provided in Table \ref{table 1}.
\begin{table}[t]
\centering
\caption{Example of decision matrices-subjective attributes}
\makebox[ 0.55\textwidth][c]{       
\resizebox{0.55 \textwidth}{!}{ 
\begin{tabular}{@{\extracolsep{4pt}}c c c c c c c c c  @{}}
\hline
\multirow{2}{*}{No. of Candidates} &  \multicolumn{2}{c}{$u_{1}$} & \multicolumn{2}{c}{$u_{2}$} &  \multicolumn{2}{c}{$u_{3}$} & \multicolumn{2}{c}{$u_{4}$}\\
\cline{2-3}
\cline{4-5}
\cline{6-7}
\cline{8-9}
 & \makebox[2cm]{\shortstack{Panel\\interview}} & \makebox[2cm]{\shortstack{1-on-1\\interview}} 
 & \makebox[2cm]{\shortstack{Panel\\interview}} & \makebox[2cm]{\shortstack{1-on-1\\interview}} 
 & \makebox[2cm]{\shortstack{Panel\\interview}} & \makebox[2cm]{\shortstack{1-on-1\\interview}} 
 & \makebox[2cm]{\shortstack{Panel\\interview}} & \makebox[2cm]{\shortstack{1-on-1\\interview}} \\
 \hline
 1 & 80 & 75 & 85 & 80 & 75 & 70 & 90 & 85\\
 2 & 65 & 75 & 60 & 70 & 70 & 77 & 60 & 70\\
 3 & 90 & 85 & 80 & 85 & 80 & 90 & 90 & 95 \\
 4 & 65 & 70 & 55 & 60 & 68 & 72 & 62 & 72\\
 5 & 75 & 80 & 75 & 80 & 50 & 55 & 70 & 75\\
 6 & 80 & 80 & 75 & 85 & 77 & 82 & 75 & 75\\
 7 & 65 & 70 & 70 & 60 & 65 & 72 & 67 & 75\\
 8 & 70 & 60 & 75 & 65 & 75 & 67 & 82 & 85\\
 9 & 80 & 85 & 95 & 85 & 90 & 85 & 90 & 92\\
 10 & 70 & 75 & 75 & 80 & 68 & 78 & 65 & 70\\
 11 & 50 & 60 & 62 & 65 & 60 & 65 & 65 & 70\\
 12 & 60 & 65 & 65 & 75 & 50 & 60 & 45 & 50\\
 13 & 75 & 75 & 80 & 80 & 65 & 75 & 70 & 75\\
 14 & 80 & 70 & 75 & 72 & 80 & 70 & 75 & 75\\
 15 & 70 & 65 & 75 & 70 & 65 & 70 & 60 & 65\\
 16 & 90 & 95 & 92 & 90 & 85 & 80 & 88 & 90\\
 17 & 80 & 85 & 70 & 75 & 75 & 80 & 70 & 75\\
 \hline
\end{tabular}
}}

\label{table 1}
\end{table}
\newline
\textbf{Step 1.} Construct the normalized decision matrices for Table \ref{table 1}, followed by BPA generation for each alternative corresponding to each attribute with the respective expert. In this study, we have determined a $5$ linguistic term set $\{verylow, low, medium, high, veryhigh\}$ and corresponding triangular membership functions $\{\mu_{\tilde{B_{1}}},\mu_{\tilde{B_{2}}},...,\mu_{\tilde{B_{5}}}\}$ to characterize the alternative's belongingness with respect to the attributes. According to the empirical investigation performed by psychologist Miller \cite{miller1956magical}, the utilization of less than five linguistic terms is not competent for apprehending sufficient information. Selecting more than 9 linguistic terms is disproportionate to understanding the essential differences. Therefore, a set of $7\pm 2$ linguistic terms is needed for characterizing the objectives and decision variables in real decision situations \cite{miller1956magical}. 
The evaluation of the membership function followed by the BPAs is illustrated in Table \ref{table 3} and \ref{table 4} using a set of $5$ linguistic term sets.
\begin{table*}[!t]
\centering
\caption{Membership of each alternative with respect to the attributes for expert $u_{1}$}
\makebox[ 0.7\textwidth][c]{       
\resizebox{0.7 \textwidth}{!}{ 
\begin{tabular}{@{\extracolsep{4pt}}c c c c c c c c c c c @{}}
\hline
\multirow{2}{*}{No. of Candidates} &  \multicolumn{5}{c}{Panel interview} & \multicolumn{5}{c}{1-on-1 interview} \\
\cline{2-6}
\cline{7-11}
 & $\mu^{u_{1}}_{t_{1}\Tilde{B}_{1}}$ & $\mu^{u_{1}}_{t_{1}\Tilde{B}_{2}}$ & $\mu^{u_{1}}_{t_{1}\Tilde{B}_{3}}$ 
 & $\mu^{u_{1}}_{t_{1}\Tilde{B}_{4}}$ & $\mu^{u_{1}}_{t_{1}\Tilde{B}_{5}}$ 
 & $\mu^{u_{1}}_{t_{2}\Tilde{B}_{1}}$ & $\mu^{u_{1}}_{t_{2}\Tilde{B}_{2}}$ & $\mu^{u_{1}}_{t_{2}\Tilde{B}_{3}}$ 
 & $\mu^{u_{1}}_{t_{2}\Tilde{B}_{4}}$ & $\mu^{u_{1}}_{t_{2}\Tilde{B}_{5}}$\\ 
 \hline
 1 & 0.2500 & 0.3333 & 0.5000 & 1.000 & 0.7500 & 0.5714 & 0.7619 & 0.8571 & 0.5714 & 0.4286 \\
2 & 0.6250 & 0.8333 & 0.7500 & 0.5000 & 0.3750 & 0.5714 & 0.7619 & 0.8571 & 0.5714 & 0.4286 \\
 3 & 0.0000 & 0.0000 & 0.0000 & 0.0000 & 1.0000 & 0.2857 & 0.3809 & 0.5714 & 0.9523 & 0.7142 \\
4 & 0.6250 & 0.8333 & 0.7500 & 0.5000 & 0.3750 & 0.7142 & 0.9523 & 0.5714 & 0.3809 & 0.2857 \\
 5 & 0.3750 & 0.5000 & 0.7500 & 0.8333 & 0.6250 & 0.4285 & 0.5714 & 0.8571 & 0.7619 & 0.5714 \\
 6 & 0.2500 & 0.3333 & 0.5000 & 1.0000 & 0.7500 & 0.4285 & 0.5714 & 0.8571 & 0.7619 & 0.5714 \\
 7 & 0.6250 & 0.8333 & 0.7500 & 0.5000 & 0.3750 & 0.7143 & 0.9523 & 0.5714 & 0.3809 & 0.2857 \\
 8 & 0.5000 & 0.6667 & 1.0000 & 0.6667 & 0.5000 & 1.0000 & 0.0000 & 0.0000 & 0.0000 & 0.0000\\
 9 & 0.2500 & 0.3333 & 0.5000 & 1.0000 & 0.7500 & 0.2857 & 0.3809 & 0.5714 & 0.9523 & 0.7143 \\
 10 & 0.5000 & 0.6667 & 1.0000 & 0.667 & 0.5000 & 0.5714 & 0.7619 & 0.8571 & 0.5714 & 0.4286 \\
 11 & 1.0000 & 0.0000 & 0.0000 & 0.0000 & 0.0000 &  1.0000 & 0.0000 & 0.0000 & 0.0000 & 0.0000 \\
 12 & 0.7500 & 1.0000 & 0.5000 & 0.3333 & 0.2500 & 0.8571 & 0.5714 & 0.2857 & 0.1904 & 0.1428 \\
 13 & 0.3750 & 0.5000 & 0.7500 & 0.8333 & 0.6250 & 0.5714 & 0.7619 & 0.8571 & 0.5714 & 0.3809 \\
 14 & 0.2500 & 0.3333 & 0.5000 & 1.0000 &  0.7500 & 0.7142 & 0.9523 & 0.5714 & 0.3809 & 0.2857 \\
 15 & 0.5000 & 0.6667 & 1.0000 &  0.6667 & 0.5000 & 0.8571 & 0.5714 & 0.2857 & 0.1905 & 0.1428 \\
 16 & 0.0000 & 0.0000 & 0.0000  & 0.0000 & 1.0000 & 0.0000 & 0.0000 & 0.0000 & 0.0000 & 1.0000 \\
 17 & 0.2500 & 0.3333 & 0.5000 & 1.0000 & 0.7500 & 0.2857 & 0.3809 & 0.5714 & 0.9523& 0.7143\\
\hline
\end{tabular}
}}
\label{table 3}
\end{table*}
\begin{table*}[!t]
\centering
\caption{BPAs of each alternative with respect to the attributes for expert $u_{1}$}
\makebox[ 0.7\textwidth][c]{       
\resizebox{0.7 \textwidth}{!}{ 
\begin{tabular}{@{\extracolsep{4pt}}c c c c c c c c c c c @{}}
\hline
\multirow{2}{*}{No. of Candidates} &  \multicolumn{5}{c}{Panel interview} & \multicolumn{5}{c}{1-on-1 interview} \\
\cline{2-6}
\cline{7-11}
 & $m^{u_{1}}_{t_{1}\Tilde{B}_{1}}$ & $m^{u_{1}}_{t_{1}\Tilde{B}_{2}}$ & $m^{u_{1}}_{t_{1}\Tilde{B}_{3}}$ 
 & $m^{u_{1}}_{t_{1}\Tilde{B}_{4}}$ & $m^{u_{1}}_{t_{1}\Tilde{B}_{5}}$ 
 & $m^{u_{1}}_{t_{2}\Tilde{B}_{1}}$ & $m^{u_{1}}_{t_{2}\Tilde{B}_{2}}$ & $m^{u_{1}}_{t_{2}\Tilde{B}_{3}}$ 
 & $m^{u_{1}}_{t_{2}\Tilde{B}_{4}}$ & $m^{u_{1}}_{t_{2}\Tilde{B}_{5}}$\\ 
 \hline
 
 1 & 0.0351 & 0.0408 & 0.0513 & 0.0952 & 0.0759 & 0.0579 & 0.0816 & 0.0937 & 0.0697 & 0.0600\\
2 & 0.0877 & 0.1021 & 0.0769 & 0.0476 & 0.0379 & 0.0579 & 0.0816 & 0.0937 & 0.0697 & 0.0600 \\
 3 & 0.0000 & 0.0000 & 0.0000 & 0.0000 & 0.1012 & 0.0289 & 0.0408 & 0.0625 & 0.1163 & 0.1000 \\
4 & 0.0877 & 0.1021 & 0.0769 & 0.0476 & 0.0379 & 0.0725 & 0.1021 & 0.0625 & 0.0465 & 0.0400 \\
 5 & 0.0526 & 0.0612 & 0.0769 & 0.0793 & 0.0633 & 0.0435 & 0.0612 & 0.0937 & 0.0931 & 0.0800 \\
 6 & 0.0351 & 0.0408 & 0.0512 & 0.0952 & 0.0759 & 0.0435 & 0.0612 & 0.0937 & 0.0931 & 0.0800 \\
 7 & 0.0877 & 0.1021 & 0.0769 & 0.0476 & 0.0379 & 0.0725 & 0.1021 & 0.0625 & 0.0465 & 0.0400 \\
 8 & 0.0702 & 0.0816 & 0.1026 & 0.0635 & 0.0506 & 0.1014 & 0.0000 & 0.0000 & 0.0000 & 0.0000\\
 9 & 0.0351 & 0.0408 & 0.0512 & 0.0952 & 0.0759 & 0.0289 & 0.0408 & 0.0625 & 0.1162 & 0.1000 \\
 10 & 0.0702 & 0.0816 & 0.1026 & 0.0635 & 0.0506 & 0.0579 & 0.0816 & 0.0938 & 0.0697 & 0.0600 \\
 11 & 0.1404 & 0.0000 & 0.0000 & 0.0000 & 0.0000 &  0.1015 & 0.0000 & 0.0000 & 0.0000 & 0.0000 \\
 12 & 0.1053 & 0.1224 & 0.0513 & 0.0317 & 0.0253 & 0.0869 & 0.0612 & 0.0313 & 0.0233 & 0.0200 \\
 13 & 0.0526 & 0.0612 & 0.0769 & 0.0794 & 0.0633 & 0.0579 & 0.0816 & 0.0938 & 0.0698 & 0.0600 \\
 14 & 0.0351 & 0.0408 & 0.0513 & 0.0952 &  0.0759 & 0.0724 & 0.1021 & 0.0625 & 0.0465 & 0.0400 \\
 15 & 0.0702 & 0.0816 & 0.1026 &  0.0635 & 0.0506 & 0.0869 & 0.0612 & 0.0313 & 0.0233 & 0.0200 \\
 16 & 0.0000 & 0.0000 & 0.0000  & 0.0000 & 0.1013 & 0.0000 & 0.0000 & 0.0000 & 0.0000 & 0.1400\\
 17 & 0.0351 & 0.0408 & 0.0512 & 0.0952 & 0.0759 & 0.0289 & 0.0408 & 0.0625 & 0.1163 & 0.1000\\
\hline
\end{tabular}
}}

\label{table 4}
\end{table*}
\\
\textbf{Step 2.} Determine the ordered weighted belief measure for each alternative corresponding to the experts with respect to the attributes.
\\
\textbf{Step 3.} Determine the ordered weighted plausibility measure for each alternative corresponding to the experts with respect to the attributes to handle the conflicts and impreciseness occurring between the group of experts.
\newline
\textbf{Step 4.} Determine $WPBl_{w1}, WPBl_{w_{2}}, WPBl_{w3}, WPBl_{w4}$ using the belief and plausibility measure evaluated in Step 2 and Step 3. 
\textbf{Step 5.}
Determine the weighted belief divergence measure between each pair of a group of experts based on attributes proposition for each alternative using Step 4, and the results are reported in Table \ref{table 7}. 
\begin{table}[!t]
\centering
\caption{The proposed weighted divergence measure between each pair of a group of experts.}
\makebox[ 0.5\textwidth][c]{       
\resizebox{0.5 \textwidth}{!}{ 
\begin{tabular}{@{\extracolsep{4pt}}c c c c c c c @{}}
\hline
\multirow{1}{*}{No. of Candidates} &  \multicolumn{1}{c}{$D(u_{1},u_{2})$} & \multicolumn{1}{c}{$D(u_{1},u_{3})$} &  \multicolumn{1}{c}{$D(u_{1},u_{4})$} & \multicolumn{1}{c}{$D(u_{2},u_{3})$} & \multicolumn{1}{c}{$D(u_{2},u_{4})$} & \multicolumn{1}{c}{$D(u_{3},u_{4})$}\\
\hline
1 & 0.0006 & 0.0000 & 0.0002 & 0.0007 & 0.0016 & 0.0003 \\
2 & 0.0031 & 0.0004 & 0.0009 & 0.0059 & 0.0074 & 0.0001 \\
3 & 0.0036 & 0.0058 & 0.0029 & 0.0003 & 0.0001 & 0.0005 \\
4 & 0.0003 & 0.0001 & 0.0009 & 0.0006 & 0.0001 & 0.0013 \\
5 & 0.0004 & 0.0009 & 0.0049 & 0.0001 & 0.0026 & 0.0016 \\
6 & 0.0000 & 0.0001 & 0.0032 & 0.0001 & 0.0034 & 0.0026 \\
7 & 0.0000 & 0.0001 & 0.0020 & 0.0000 & 0.0022 & 0.0004 \\
8 & 0.0005 & 0.0034 & 0.0015 & 0.0012 & 0.0002 & 0.0004 \\
9 & 0.0039 & 0.0037 & 0.0017 & 0.0001 & 0.0005 & 0.0004 \\
10 & 0.0011 & 0.0001 & 0.0000 & 0.0009 & 0.0010 & 0.0000 \\
11 & 0.0036 & 0.0025 & 0.0029 & 0.0001 & 0.0001 & 0.0000 \\
12 & 0.0093 & 0.0053 & 0.0045 & 0.0006 & 0.0009 & 0.0000 \\
13 & 0.0004 & 0.0040 & 0.0044 & 0.0019 & 0.0021 & 0.0001 \\
14 & 0.0002 & 0.0021 & 0.0041 & 0.0035 & 0.0060 & 0.0003 \\
15 & 0.0065 & 0.0003 & 0.0008 & 0.0041 & 0.0028 & 0.0001 \\
16 & 0.0045 & 0.0044 & 0.0045 & 0.0000 & 0.0000 & 0.0000 \\
17 & 0.0001 & 0.0007 & 0.0061 & 0.0014 & 0.0081 & 0.0027 \\
\hline
Average & 0.0023 & 0.0021 & 0.0027 & 0.0012 & 0.0023 & 0.0007\\
\hline
\end{tabular}
}}
\label{table 7}
\end{table}
\newline
\textbf{Step 6.} Construct the divergence measure matrix to handle the impreciseness and to consider inter-observational variability between each group of experts corresponding to the attributes for each alternative. The divergence measure matrix $D_{MM}=[d_{ij}]_{4 \times 4}$ is evaluated as follows:
 \begin{equation*}
     D_{MM} = \bordermatrix{ & u_{1} & u_{2} & u_{3}  & u_{4} \cr
               u_{1} & 0.0000 & 0.0023 & 0.0021 & 0.0027 \cr
               u_{2} & 0.0021 & 0.0000 & 0.0012 & 0.0023 \cr
                u_{3} & 0.0021 & 0.0012 & 0.0000 & 0.0007 \cr 
               u_{4} & 0.0027 & 0.0023 & 0.0007 & 0.0000 } 
 \end{equation*}
\textbf{Step 7.} Calculate the average divergence measure corresponding to the $D_{MM}$ for each experts $u_{k}$ as follows: $\Tilde{D}_{u_{1}} = 0.0017, \quad \Tilde{D}_{u_{2}} = 0.0014, \quad \Tilde{D}_{u_{3}} = 0.0010, \quad \Tilde{D}_{u_{4}} = 0.0014.$
\textbf{Step 8.} Calculate the weighted support ($\Tilde{S}$) for each exert as follows: $\Tilde{S}_{u_{1}} = 573.03, \quad  \Tilde{S}_{u_{2}} = 691.65,
    \Tilde{S}_{u_{3}} = 997.56, \quad \Tilde{S}_{u_{4}} = 704.38.$
\\    
\textbf{Step 9.} Calculate the final weighted support for each expert as follows: $W_{u_{1}}=0.1932, \quad W_{u_{2}}= 0.2331,$ $\quad W_{u_{3}}=0.3362, \quad W_{u_{4}}=0.2374$.
The final weighted support is shown in Table \ref{table 8}, and the higher weight corresponding to the group of experts indicates the influence of attributes in group decision-making. Therefore, the final ranking of the experts is  $u_{3}\succ u_{4} \succ u_{2} \succ u_{1}$. We have compared the expert ranking with the existing methods, and the results are reported in Table \ref{table 8}.
\begin{table}[!t]
\centering
\caption{Experts weights, ranking and comparison of the results with other methods}
\makebox[ 0.5\textwidth][c]{       
\resizebox{0.5 \textwidth}{!}{ 
\begin{tabular}{@{\extracolsep{4pt}}l c c @{}}
\hline
\multirow{1}{*}{Methods} &  \multicolumn{1}{c}{Weights of the experts} & \multicolumn{1}{c}{Ranking orders}\\
\hline
Z.Yue \cite{yue2011method} & $\lambda_{1}=0.2350, \lambda_{2}=0.2601, \lambda_{3}=0.2485, \lambda_{4}=0.2564$ & $u_{2}\succ u_{4} \succ u_{3} \succ u_{1}$\\
Z.Yue \cite{yue2012approach} & $\lambda_{1}=0.2478, \lambda_{2}=0.2502, \lambda_{3}=0.2517, \lambda_{4}=0.2503$ & $u_{3}\succ u_{4} \succ u_{2} \succ u_{1}$\\
Proposed (Evidential MAGDM) & $W_{u_{1}}=0.1932, W_{u_{2}}=0.2331, W_{u_{3}}=0.3362, W_{u_{4}}=0.2374$ & $u_{3}\succ u_{4} \succ u_{2} \succ u_{1}$\\
\hline
\end{tabular}
}}
\label{table 8}
\end{table}
\begin{table}[!t]
\centering
\caption{Collective analysis of all the alternatives.}
\makebox[ 0.5\textwidth][c]{       
\resizebox{0.5 \textwidth}{!}{ 
\begin{tabular}{@{\extracolsep{4pt}}c c c c c c c c@{}}
\hline
\multirow{1}{*}{No. of Candidates} &  \multicolumn{1}{c}{Panel interview} & \multicolumn{1}{c}{1-on-1 interview} &  \multicolumn{1}{c}{Weights} & \multicolumn{1}{c}{Ranking} & \multicolumn{1}{c}{Shih et al. \cite{shih2007extension}} & \multicolumn{1}{c}{Yue \cite{yue2012approach}} & \multicolumn{1}{c}{}\\
\hline
1 & 0.2715 & 0.2474 & 0.4809 & 4 & 5 & 4 & \\
2 & 0.2137 & 0.2364 & 0.4166 & 12 & 14 & 12 &\\
3 & 0.2797 & 0.2869 & 0.5247 & 3 & 3 & 3 &\\
4 & 0.2093 & 0.2218 & 0.3991 & 15 & 12 & 15 &\\
5 & 0.2164 & 0.2264 & 0.4101 & 13 & 11 & 11 &\\
6 & 0.2544 & 0.2599 & 0.4763 & 5 & 4 & 5 &\\
7 & 0.2211 & 0.2241 & 0.4122 & 12 & 13 & 13 &\\
8 & 0.2513 & 0.2235 & 0.4402 & 9 & 8 & 10 &\\
9 & 0.2961 & 0.2791 & 0.5331 & 1 & 2 & 2 &\\
10 & 0.2299 & 0.2449 & 0.4396 & 10 & 10 & 9 &\\ 
11 & 0.1982 & 0.2101 & 0.3781 & 16 & 16 & 16 &\\
12 & 0.1796 & 0.2002 & 0.3515 & 17 & 17 & 17 &\\
13 & 0.2373 & 0.2454 & 0.4470 & 8 & 9 & 7 &\\
14 & 0.2578 & 0.2308 & 0.4530 & 7 & 6 & 8 &\\
15 & 0.2225 & 0.2187 & 0.4087 & 14 & 15 & 14  &\\
16 & 0.2929 & 0.2821 & 0.5328 & 2 & 1 & 1 &\\
17 & 0.2444 & 0.2534 & 0.4609 & 6 & 7 & 6 &\\
\hline
Ideal Solution & 0.2961 & 0.2869 & - & - & - & - & -\\
\hline
\end{tabular}
}}

\label{table 9}
\end{table}
\newline
\textbf{Step 10.} Perform the fusion of attributes corresponding to the experts by providing weightage to each attribute determined in Step 9 and rank as shown in Table \ref{table 9}. The performance of the Evidential MAGDM is compared with the existing method illustrated in Table \ref{table 9}, which shows the effectiveness of the proposed method when expert weightage is not included, whereas it is incorporated in the existing approaches.\\
Next, we show the application of the proposed Evidential MAGDM for the ensemble classifier feature fusion for the classification of retinal disorders using OCT images.
\raggedbottom
\subsection{Ensemble classifier feature fusion for classification using the proposed Evidential MAGDM method}{\label{sec 4.2}}
This section introduces the real-world application of the proposed Evidential MAGDM
framework in the ensemble classifier feature fusion for the diagnosis of retinal disorders using OCT images illustrated in Fig. \ref{fig 3}. To analyze the effectiveness of the proposed Evidential MAGDM for feature fusion, we have employed a publicly available \cite{kulyabin2024octdl} OCT image dataset and compared it with state-of-the-art methods. Additionally, the experimental performance of the proposed method is compared by taking distinct weights for fusing features. 
\subsubsection{Proposed Ensemble Feature Fusion Classifier for Diagnosis of Retinal Disorders}
\begin{figure*}[!t]
   \centering
    \includegraphics[width=7in]{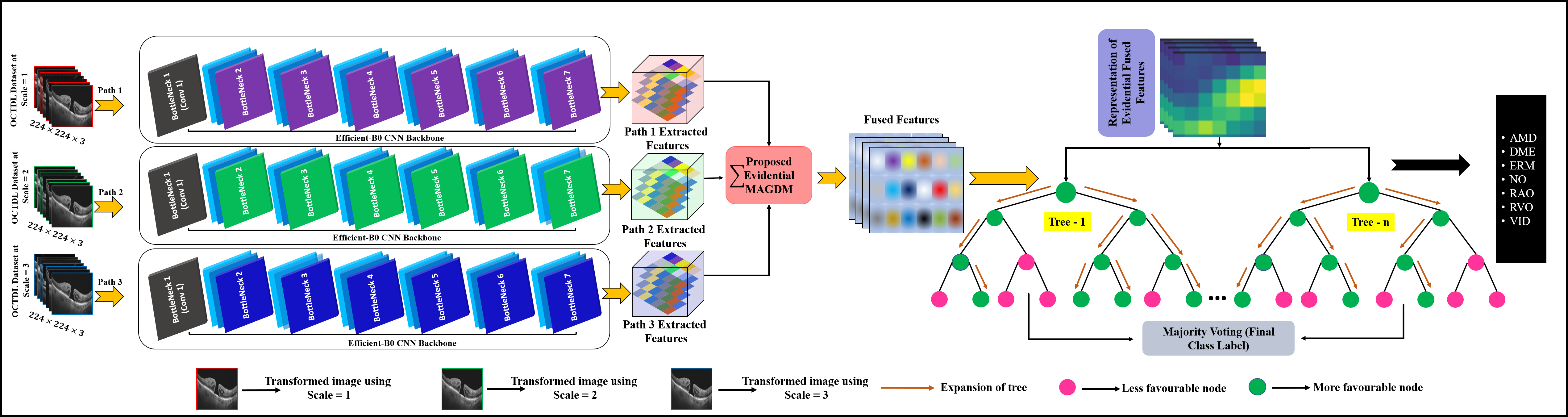}
    \caption{The proposed Evidential MAGDM feature fusion module integrated into the ensemble classifier to analyze inter-observational variability between the features.}
    \label{fig 3}
\end{figure*}

{\label{sec 4.2.1}}
Several crucial ocular and systematic disorders may cause vision loss or even blindness. Retinal imagining approaches are widely employed in ophthalmology to diagnose ocular disorders in a non-invasive manner. Optical coherence tomography (OCT) is currently the most widespread imaging modality for analyzing several retinal disorders, such as diabetic retinopathy, age-related macular degeneration, and glaucoma \cite{schmidt2018artificial,burlina2017automated}. Optical coherence tomography (OCT) provides a cross-sectional view of biological tissues at microscopic spatial resolution level \cite{van2007recent} and the surface information of the retinal layers, which has a major contribution to the early diagnosis of ocular disease. However, the interpretation of 3D OCT images is a time-consuming procedure for the ophthalmologist. Therefore, various automated computer-aided diagnoses have been introduced during recent years \cite{sunija2021octnet,george2024two,al2024fundus,elsharkawy2024clinically} to analyze OCT data. 
\\
We propose a multi-scale space problem by taking different regularization levels to control the smoothing of the OCT image dataset, which functions as different experts illustrated in Fig. \ref{fig 3}. We assess three scale-spaces that induce multiple sets of OCT image datasets (considered as experts) and feed them to three homogeneous EfficientNetB0 \cite{tan2019efficientnet} models for extracting the features that are assumed as attributes, and the spatial resolution of features are considered as alternatives. The extracted features through the considered model EfficientNetB0 \cite{tan2019efficientnet} characterize variations in the discriminative and textural features at multi-scale. The representation of extracted features from each path is illustrated in Fig. \ref{fig 3}. Now, these extracted features are fused by incorporating the proposed Evidential MAGDM fusion module.  For each feature space, the weights are generated by analyzing the inter-observational variability between the extracted features through the proposed Evidential MAGDM approach, which also handles the impreciseness emerging between the distinct scale spaces. Finally, these fused features work as input for the random forest classifier (RFC) model using the transfer learning mechanism for classification purposes. The Evidential MAGDM method apprehends the variations in the features during the fusion procedure by optimizing the impreciseness of the multi-scale features and computes the weights in such a way that the performance of the model is not biased by suppressing the conflicts emerging in the feature information.
\subsubsection{Dataset and Implementation Details}
\begin{figure*}[t]
   \centering
    \includegraphics[width=7in]{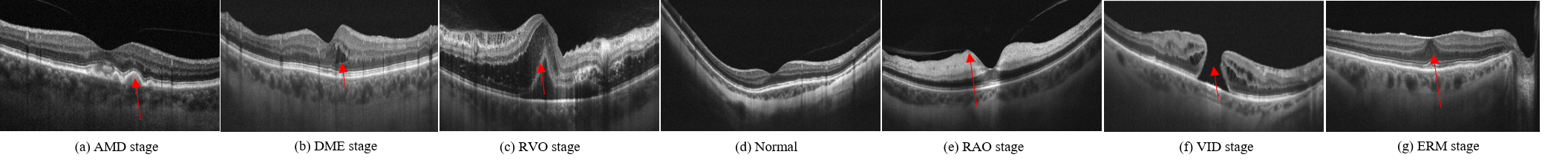}
    \caption{ Description of various retinal disorders includes in the OCTDL dataset.}
    \label{fig 4}
\end{figure*}

{\label{sec 4.2.2}}
To perform the empirical study, we have used the publicly available OCTDL dataset \cite{kulyabin2024octdl}, which includes 2064 images classified into eclectic diseases and retinal disorders. The datasets comprise high spatial resolution OCT B-scans, which enable the visualization of sub-surface layer information centered on the fovea and the choroidal blood vessel. The description of the dataset is illustrated in Table \ref{table 10}, which includes seven various classes, and it can be visualized from Fig. \ref{fig 4}.
\begin{table}[!t]
\centering
\caption{OCTDL dataset distribution corresponding to the retinal disorder}
\makebox[ 0.4\textwidth][c]{       
\resizebox{0.4 \textwidth}{!}{ 
\begin{tabular}{@{\extracolsep{4pt}}l c c  @{}}
\hline
\multirow{1}{*}{OCTDL Dataset (Retinal Disorder} &  \multicolumn{1}{c}{Number of Scans} & \multicolumn{1}{c}{Class} \\
\hline
Normal & 332 & NO \\
Age-related macular degeneration & 1231 & AMD \\
Diabetic macular edema & 147 & DME \\
Epiretinal membrane & 155 & ERM \\
Retinal artery occlusion & 22 & RAO \\
Retinal vein occlusion & 101 & RVO \\
Vitreomacular interface disease & 76 & VID\\
Total & 2064 & -\\
\hline
\end{tabular}
}}
\label{table 10}
\end{table}
The dataset is divided into 80:20 for the training and testing part. The data augmentation strategy is utilized, which retrieved 1200 images for each class of the OCTDL dataset, and the images are resized of spatial resolution of $224 \times 224$. The model is trained over the training dataset for 100 epochs with a cosine learning rate for feature extraction purposes followed by integrating the proposed Evidential MAGDM feature fusion model and fed to the RFC for the classification.
To analyze the performance of proposed Evidential MAGDM in the ensemble classifier feature fusion, we have computed the seven evaluation measures, which include Accuracy, Sensitivity, Specificity, Precision, F$_{1}$ score, AUC (area under the receiver operating characteristic), and Kappa. It can be formulated as follows: 
\small{
\begin{equation*}
Accuracy = \dfrac{TP+TN}{TP+FP+TN+FN} \quad Sensitivity = \dfrac{TP}{TP+FN} 
\end{equation*}}
\begin{equation*}
    Specificity=\dfrac{TN}{TN+FP}    \quad Precision = \dfrac{TP}{TP+FP}
\end{equation*}
\begin{equation*}
    F_{1}=\dfrac{2\times Precision\times Sensitivity}{Precision+Sensitivity}
\end{equation*}
\subsubsection{Experimental Results}{\label{sec 4.2.3}}
We investigate the efficacy of the proposed Evidential MAGDM module by taking different combinations of weights for feature fusion in the proposed model illustrated in Fig. \ref{fig 3} for the OCTDL dataset. The comparison is accomplished with the same network backbone(EfficientB0), including training and testing strategies. We have considered a set of distinct weights to analyze the effectiveness of the proposed method. The experimental performance is reported in Table \ref{table 11} in terms of evaluation measures presented in section \ref{sec 4.2.2} over the OCTDL dataset for the classification of diverse retinal disorders in OCT images. The evaluation measures for the proposed Evidential MAGDM feature fusion module are presented in blue color in Table \ref{table 11}. 
\begin{table}[!t]
\centering
\caption{Average evaluation measures over the OCTDL datasets with various considered weights for fusing the extracted features and comapred with the state-of-the-art methods}
\makebox[ 0.5\textwidth][c]{       
\resizebox{0.5 \textwidth}{!}{ 
\begin{tabular}{@{\extracolsep{4pt}}l c c c c c c c@{}}
\hline
\multirow{1}{*}{Method} &  \multicolumn{1}{c}{Accuracy} & \multicolumn{1}{c}{Sensitivity} & \multicolumn{1}{c}{Specificity}  & \multicolumn{1}{c}{Precision}
& \multicolumn{1}{c}{F$_{1}$} & \multicolumn{1}{c}{AUC} & \multicolumn{1}{c}{Kappa}  \\
\hline

EfficientB0$_{(1,0,0)}$ & 0.877 & 0.830 & 0.976 & 0.836 & 0.880 & 0.956 & 0.802 \\
EfficientB0$_{(0,1,0)}$ & 0.901 & 0.861 & 0.980 & 0.873 & 0.903 & 0.962 & 0.839\\
EfficientB0$_{(0,0,1)}$ & 0.863 & 0.763 & 0.971 & 0.814 & 0.853 & 0.961 & 0.775\\
EfficientB0$_{(0.5,0.5,0)}$ & 0.901 & 0.855 & 0.980 & 0.881 & 0.902 & 0.973 & 0.840\\
EfficientB0$_{(0.5,0.25,0.25)}$ & 0.906 & 0.868 & 0.982 & 0.883 & 0.907 & 0.977 & 0.848\\
\hline
ResNet50 \cite{kulyabin2024octdl} & 0.846 & 0.846 & \_ & 0.898 & 0.866 & \underline{0.988} & \_\\
VGG16 \cite{kulyabin2024octdl} & 0.859 & 0.859 & \_ & 0.888 & 0.869 & 0.977 & \_  \\
Sunija et al. \cite{sunija2021octnet} & 0.882 & 0.882 & 0.883 & 0.884 & 0.881 & 0.971 & 0.807\\

Geroge et al.\cite{george2024two} & 0.873 & 0.873 & \_ & 0.879 & 0.874 & 0.975 & 0.795 \\

\textcolor{blue}{EfficientB0 (Evidential MAGDM)} & \textcolor{blue}{0.911} & \textcolor{blue}{0.870} & \textcolor{blue}{0.982} & \textcolor{blue}{0.901} & \textcolor{blue}{0.912} & \textcolor{blue}{0.981} & \textcolor{blue}{0.855}\\

\hline
\end{tabular}
}}
\label{table 11}
\end{table}
\begin{figure}[!t]
   \centering
    \includegraphics[width=3.3in]{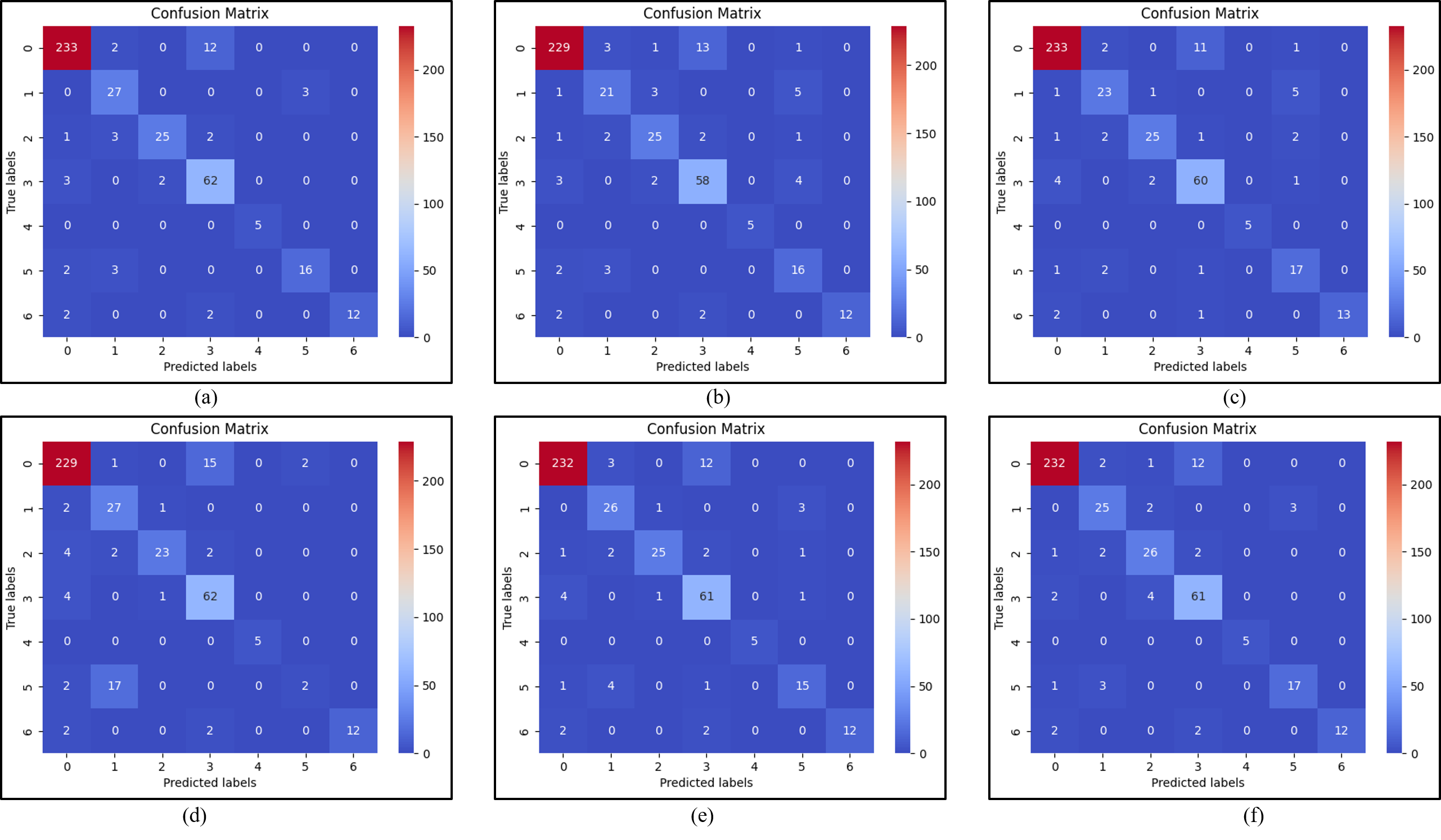}
    \caption{The confusion matrix of the proposed Evidential MAGDM method in the homogeneous ensemble classifier feature fusion with different weights (a) EfficientB0 (Evidential MAGDM), (b) EfficientB0$_{(1,0,0)}$, (c) EfficientB0$_{(0,1,0)}$, (d) EfficientB0$_{(0,0,1)}$, (e) EfficientB0$_{(0.5,0.5,0)}$, (f) EfficientB0$_{(0.5,0.25,0.25)}$.}
    \label{fig 5}
\end{figure}
The proposed Evidential MAGDM module achieves 0.911 Accuracy, which is adequate compared to the other considered weights. When we are considering one expert for decision (classification), the performance of the model is reduced compared to the proposed Evidential MAGDM, especially for the $(1,0,0)$ and $(0,0,1)$ weights in terms of all the evaluation measures. When we are evaluating the model performance over the weight $(0,0,1)$, the Sensitivity of the model is lower compared to others, which indicates that the model is not able to capture the discriminative feature representation when conflicting and impreciseness is included, which increments the false positive. However, the performance of the model is increased for the $(0.5,0.25,0.25)$ weights compared to other considered weights while lower than the proposed strategy. It can be observed that taking random weights for the experts corresponding to the feature fusion leads to degradation of the model's performance, and choosing the optimal weight for feature fusion is tedious. The confusion matrix is demonstrated in Fig. \ref{fig 5} for all the considered weights and the proposed Evidential MAGDM module to show the performance of the proposed method. It can be analyzed the proposed method classification is adequate compared to other experts' weights. To analyze the behavior of the proposed Evidential MGDM in the ensemble model for classification, the ROC curve is plotted. The classification performance of the proposed method is satisfactory at different thresholds compared to other experts' weights, which indicates the effectiveness of the proposed Evidential MAGDM approach to handling the impreciseness and considering the inter-observational variability between the features extracted through different experts. The performance of the proposed method is compared with the state-of-the-art methods illustrated in Table \ref{table 11}. The proposed method achieves a higher Accuracy of 0.911, which shows the robustness of the Evidential MAGDM approach in the ensemble classifier fusion. Additionally, the proposed method achieves higher evaluation metrics compared to the \cite{kulyabin2024octdl,sunija2021octnet,george2024two}. The experimental results indicate the efficacy of the proposed Evidential MAGDM in the ensemble classifier feature fusion compared to the utilization of the single classifier. The proposed method can help in decision-making for the diagnosis of retinal disorders when various ophthalmologists are involved.  
\subsection{Discussion}{\label{sec 4.3}}
Through the experimental analysis of the proposed Evidential MAGDM approach in the illustrative example and ensemble classifier feature fusion, the effectiveness of the proposed method has been affirmed. The proposed method has some advantages over the existing MAGDM. The proposed MAGDM method can effectively handle the conflicts occurring between the group of experts, which can influence decision-making. The existing method required attribute weight information for the respective experts. However, if there is a lack of weight attribute information, then it may affect the performance of the MAGDM approach, while it can not affect the proposed method decision-making. Additionally, the ordered weighted belief computed for each alternative considers the inter-variability between the other alternatives for the respective attribute by handling the impreciseness and integrating the partitioning of the domain of the attribute into linguistic terms. The ranking results of the group of experts are illustrated in Table \ref{table 8} and compared with the existing approach. It can be noted that the lack of attribute weights does not affect the final outcome illustrated in Table \ref{table 8}. However, the attribute weights are utilized in the presented method \cite{yue2011method,yue2012approach}, which may affect the decision if there is a lack of attribute weight information. The proposed Evidential MAGDM has a vast range of applications in the medical domain. We have analyzed the effectiveness of the proposed method by integrating it in the ensemble classifier feature fusion for the diagnosis of the retinal disorder using OCT images. The performance of the proposed approach is demonstrated in Table \ref{table 11}, which indicates the effectiveness of the proposed method over the existing models. The extracted feature information from the multi-scale considers the distinct type of feature information included in the original source. Therefore, the weight determination method is chosen appropriately for fusing the feature information so that biases and imbalances will not emerge, which can be handled through the proposed method. The experimental results of the proposed method in the ensemble classifier feature fusion indicate the effectiveness of the computed weight for feature fusion. Further, it can be visualized through the confusion matrix illustrated in Fig. \ref{fig 5} that the classification performance of the proposed method is adequately compared to the other fusion methods.
\section{Conclusion}{\label{sec 5}}
In this study, we propose a novel Evidential MAGDM method by constructing the ordered weighted belief divergence measure to compute the expert weightage for MAGDM. To evaluate the ordered weighted belief and plausibility measure, we have proposed a novel generation method of BPA for computing the degree of belief for each alternative corresponding to the attributes of each expert. The proposed methodology is able to capture the inter-observational behavior occurring between the alternatives for the respective attributes, followed by handling the conflicts and impreciseness emerging between the group of experts. To analyze the effectiveness of the proposed MAGDM method, we have shown the illustrative example and ranked the preference between the alternatives. Further, we have shown the real-world application for the diagnosis of retinal disorder using OCT images by modeling ensemble classifier feature fusion based on the proposed Evidential MAGDM model. The empirical study of the proposed method is accomplished over the publicly available OCTDL dataset and compared with different experts' weights, followed by a comparison with the state-of-the-art methods. The experimental performance indicates the effectiveness of the proposed Evidential MAGDM method and it outperforms compared with others. Moreover, we try to incorporate the multimodality imaging techniques for the diagnosis of the retinal disorder and localization by focusing on the enhancement of the Evidential MAGDM approach for the feature fusion mechanism.  
\bibliographystyle{IEEEtran}
\bibliography{References}
\end{document}